\documentclass{article}

     \PassOptionsToPackage{numbers, compress}{natbib}

     \usepackage[preprint]{neurips_2023}

\usepackage[utf8]{inputenc} 
\usepackage[T1]{fontenc}    
\usepackage{url}            
\usepackage{booktabs}       
\usepackage{amsfonts}       
\usepackage{nicefrac}       
\usepackage{microtype}      
\usepackage{xcolor}         
\usepackage{graphicx} 
\usepackage{amsfonts}
\usepackage{amsmath}
\usepackage{amsthm}
\usepackage{float}
\usepackage{enumitem}
\usepackage{color,soul}
\usepackage{subcaption}
\usepackage{algorithm}
\usepackage{algpseudocode}
\usepackage{hyperref}
\newtheorem{theorem}{Theorem}
\newlist{steps}{enumerate}{1}
\setlist[steps, 1]{label = Step \arabic*:}

\title{HLoOP $-$ Hyperbolic 2-space Local Outlier Probabilities}

\author{%
\vspace{0.2cm}
\normalfont Clémence Allietta$^{1}$\,\,\,\,\,\, Jean-Philippe Condomines$^{1}$\,\,\,\,\,\,  Jean-Yves Tourneret$^{2}$\,\,\,\,\,\, Emmanuel Lochin$^{1}$\\
\vspace{0.05cm}
  ENAC$^{1}$, IRIT$^{2}$, Université de Toulouse, France\\
  \texttt{\{name.surname\}@enac.fr}$^{1}$,\,\, \texttt{\{jean-yves.tourneret\}@irit.fr}$^{2}$ 
}

\begin{document}

\maketitle

\begin{abstract}

Hyperbolic geometry has recently garnered considerable attention in machine learning due to its capacity to embed hierarchical graph structures with low distortions for further downstream processing. This paper introduces a simple framework to detect local outliers for datasets grounded in hyperbolic 2-space referred to as HLoOP (Hyperbolic Local Outlier Probability). Within a Euclidean space, well-known techniques for local outlier detection are based on the Local Outlier Factor (LOF) and its variant, the LoOP (Local Outlier Probability), which incorporates probabilistic concepts to model the outlier level of a data vector. The developed HLoOP combines the idea of finding nearest neighbors, density-based outlier scoring with a probabilistic, statistically oriented approach. Therefore, the method consists in computing the Riemmanian distance of a data point to its nearest neighbors following a Gaussian probability density function expressed in a hyperbolic space. This is achieved by defining a Gaussian cumulative distribution in this space.
The HLoOP algorithm is tested on the WordNet dataset yielding promising results. Code and data will be made available on request for reproductibility.

\end{abstract}
\section{Introduction and Prior Work}
From social interaction analysis in social sciences to sensor networks in communication, machine learning has gained in importance in the last few years for analyzing large and complex datasets. Applying machine learning algorithms in an Euclidean space is efficient when data have an underlying Euclidean structure. However, in many applications such as computer graphics or computer vision, data cannot be embedded in a Euclidean space, which prevents the use of conventional algorithms \cite{insufisance_euclidean_space}. As an example, in datasets having a hierarchical structure, the number of relevant features can grow exponentially with the depth of the hierarchy and thus these features cannot be embedded without distortions in an Euclidean space. 
In the quest for a more appropriate geometry of hierarchies, hyperbolic spaces and their models (Poincaré disk or upper-half plane conformal models, Klein non-conformal model, Beltrami hemisphere model and Lorentz hyperboloid model among others \cite{Anderson:2006aa}) provide attractive properties that can lead to substantial performance and efficiency benefits for learning representations of hierarchical and graph data. 
Among several potential advantages, we can highlight \cite{Peng:2022aa} 1) a better generalization capability of the model, with less overfitting, computational complexity, and requirement of training data; 2) a reduction in the number of model parameters and embedding dimensions; 3) a better model understanding and interpretation. Empowered by these geometric properties, hierarchical embeddings have recently been investigated \cite{Nickel:2017aa} for complex trees with low distortions \cite{Sarkar:2011aa,bourgain,GaneaO:2018aa}. This has led to rapid advances in machine learning and data science across many disciplines and research areas, including but not limited to graph networks \cite{Liu-Q:2019aa,Chami:2019aa,Dai-J:2021aa,Cetin:2022aa}, computer vision \cite{Ghadimi-Atigh-M:2022aa,Gao-Z:2021aa,Suris-D:2021aa,Dengxiong-X:2023aa,Caron-M:2021aa}, network topology analysis \cite{Kleinberg2007,Cvetkovski2009,Cassagnes2011,Lv2020}, quantum science \cite{Higgott:2021aa,Higgott:2023aa}. Finally, it is interesting to mention the recent boom in hyperbolic neural networks and hyperbolic computer vision, which is for instance reported in the recent reviews \cite{Mettes:2023aa,Peng:2022aa}. 

Motivated by these recent advances, identifying and dealing with outliers is crucial for generating trustworthy insights and making data-driven decisions in hyperbolic spaces, e.g., providing information about which nodes are highly connected (and hence more central) or which nodes correspond to outliers such that embedding methods can realistically be used to model real complex patterns.
\begin{figure}
    \centering
    \includegraphics[scale = 1.09]{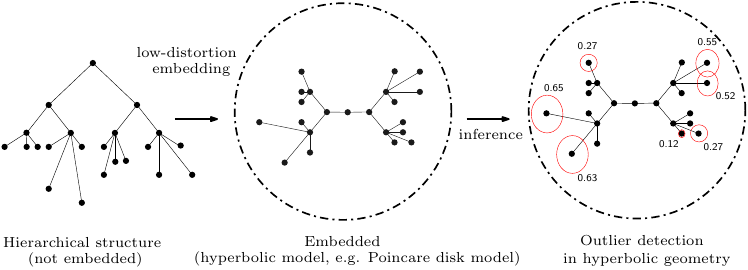}
    \vspace{0.3cm}
    \caption{Illustration inspired from \cite{Nielsen:2022aa} of local outlier probabilities in a hyperbolic model: A hierarchical structure (left) is embedded in a hyperbolic space with low-distortion (middle). The point dataset is then described by \textit{local} outlier probabilities in this hyperbolic space (right).}
    \label{fig:Hloop}
\end{figure}
In this study, we focus on \textit{local} outlier detection, which describes \textit{local} properties of data, which is relevant in many applications involving Euclidean spaces. An overview of \textit{local} anomaly detection methods can be found in the literature from many surveys or books (\cite{Su:2019aa,Alghushairy:2020aa,Souiden:2016aa,Campos:2016aa}). Initially, research related to local outlier detection was focused on intrusion detection, fraud detection\cite{Wang:2010aa}  and medical applications\cite{Lin:2005aa}. Intrusion detection \cite{Portnoy:2000aa, Garcia-Teodoro:2009aa, Yeung:2003aa} consists of detecting abnormal traffic in networks due to suspicious data or violations of network management policies. Fraud detection\cite{Phua:2010aa,Thiprungsri:2011aa,Bolton:2002aa} detects unexpected activities in banking or insurance data, such as fraudulent online payments by credit card or inconsistent insurance claims. In the wake of other disciplines, local outlier detection algorithms have been used for medical data \cite{Bansal:2016aa,Lin:2005aa}, e.g., to detect abnormal QRS complexes in electrocardiograms due to certain diseases (such as premature ventricular contraction). 

A well-known technique for local outlier detection is the Local Outlier Factor (LOF) \cite{Breunig:2000aa}\cite{Guo:2023aa} and its variant the LoOP (Local Outlier Probability)\cite{Kriegel:2009aa} with probabilistic concepts allowing the outlier level of a data point to be defined. The properties and capabilities of these methods make the detection of historical data attractive, particularly because they provide local outlier scores based on the degree of isolation of each vector from the neighborhood. While the LOF detects the outlier data points using the score of an outlier, the LoOP detects them by providing for each data point $p$ an outlier score (belonging to the interval $]0,1[$) corresponding to the probability that $p$ is an anomaly. Because the distances have positive values, the LoOP algorithm assumes a \textit{half-Gaussian} distribution for these distances. Based on Bayesian inference, the outlier score is directly interpretable as an outlier probability. 

Probabilistic inference for data embedding in hyperbolic spaces is a young research area, in which the first main contributions can be dated from the beginning of 2020 (see \cite{Nielsen:2022aa,Cho:2022aa,lorentz,Nagano:2019aa} and the references therein).
These insights led, for instance, to define the so-called Souriau Gibbs in the Poincaré disk with its Fisher information metric coinciding with the Poincaré Riemannian metric\cite{Barbaresco:2019aa}. A novel parametrization for the density of Gaussian on hyperbolic spaces is presented in \cite{Cho:2022aa}. This density can be analytically calculated and differentiated with a simple random variate generation algorithm. An alternative is to use a simple Gaussian distribution in hyperbolic spaces,e.g., \cite{Said:2014aa,Pennec:2006aa} introduced Riemannian normal distributions for the \textit{univariate normal
model}, with an application to the classification of univariate normal populations. Along with the wrapped normal generalisation used in \cite{Cho:2022aa}, \cite{Mathieu:2019aa} studies a thorough treatment of the maximum entropy normal generalisation. 
Meanwhile, a lot of applications combining hyperbolic geometry and Variational Auto-Encoders (VAEs) was investigated in \cite{poincare_random_var_dist, Mathieu:2019aa, Cho:2022ab,Nagano:2019aa} based on the fact that VAE latent space components embedded in hyperbolic space help to represent and discover hierarchies. 
This work introduces a simple framework to detect local outliers for datasets grounded in hyperbolic 2-space referred to as HLoOP (Hyperbolic Local Outlier Probability).

\paragraph{The key contributions of this paper are:} 
\begin{enumerate}[label={(\arabic*)}]
\item We extend the \textit{Local Outlier Probabilities} (LoOP) algorithm to make it applicable to  \textit{hyperbolic models}, e.g, the Poincaré disk model, leading to Hyberbolic 2-space Local Outlier probabilities (HLoOP). Figure \ref{fig:Hloop} illustrates the pipeline to obtain \textit{local} outlier probability distributions in hyperbolic geometry from hierarchical structures.
\item We derive an expression of a \textit{Gaussian cumulative distribution in hyperbolic spaces} which ensures that the Probabilistic Local Outlier Factor (PLOF) is performed by fully exploiting the information geometry of the observed data.
\end{enumerate}
The rest of this paper is structured as follows: section \ref{Univariate} briefly outlines some concepts from Riemannian geometry for univariate models. In Section \ref{Hyperbolic} we introduce the local outlier probability detection in hyperbolic spaces and discuss how this probability can be computed. Section \ref{Results} evaluates the proposed approach on the benchmark dataset `` taxonomy embedding from WordNet''.

\section{A Univariate Normal Model for Hyperbolic Spaces}\label{Univariate}
This section briefly reminds some concepts of Riemannian geometry \cite{Pennec:2006aa, Petersen:2006aa, Mathieu:2019aa} for the univariate normal model, which are necessary to formally extend the LoOP detection algorithm.

\subsection{Riemannian Geometry and Rao Distance}
A Riemannian manifold is a real and smooth manifold denoted as $\mathcal{M}$ equipped with a positive definite quadratic form $g_{\boldsymbol{x}} :\mathcal{T}_{\boldsymbol{x}}\mathcal{M}\times\mathcal{T}_{\boldsymbol{x}}\mathcal{M}\mapsto\mathbb{R}$ at each point $\boldsymbol{x}\in\mathcal{M}$, where $\mathcal{T}_{\boldsymbol{x}}\mathcal{M}$ is the tangent space defined at the local coordinates $\boldsymbol{x}=(x_1,...,x_n)^T$. Intuitively, it contains all the possible directions in which one can tangentially pass through $x$. A norm is induced by the inner product on $\mathcal{T}_{\boldsymbol{x}}\mathcal{M} : \|\cdot\|_{x}=\sqrt{<\cdot,\cdot>_{x}}$. An infinitesimal volume element is induced on each tangent space $\mathcal{T}_{\boldsymbol{x}}\mathcal{M}$. The quadratic form $g_{\boldsymbol{x}}$ is called a Riemannian metric and allows us to define the geometric properties of spaces, such as the angles and lengths of a curve. 
The Riemannian metric $g_{\boldsymbol{x}}$ is an n-by-n positive definite matrix such that an infinitesimal element of length $ds^{2}$ is defined as:
\begin{align}
 ds^{2}=\begin{pmatrix}dx_{1}\,\,\cdots\,\, dx_{n}\end{pmatrix}g_{\boldsymbol{x}}\begin{pmatrix}dx_{1}\\\,\, \vdots\,\, \\ dx_{n}\end{pmatrix}.
\end{align}
The Riemannian metric is a well-known object in differential geometry. 
For instance, the Poincaré disk with a unitary constant negative curvature corresponds to the Riemannian manifold in the hyperbolic space ($\mathbb{H}$, $g_{\boldsymbol{x}}^{\mathbb{H}})$, where $\mathbb{H}=\{\boldsymbol{x}\in\mathbb{R}^{n}:\|\boldsymbol{x}\|<1\}$ is the open unit disk\footnote{A $d$-dimensional hyperbolic space, denoted $\mathbb{H}^{d}$, is a complete, simply connected, $d$-dimensional Riemannian manifold with constant negative curvature $c$.}. Its metric tensor can be written from the Euclidean metric $g^{E}=\mathit{I}_{n}$ and the Riemannian metric such that
$g_{\boldsymbol{x}}^{\mathbb{H}}=\lambda^{2}_{x}g^{E}$, where $\lambda_{\boldsymbol{x}}=\frac{2}{1-\|\boldsymbol{x}\|^{2}}$ is the conformal factor. The Rao distance between two points $\boldsymbol{z}_{1} = (x_{1},y_{1})^T$ and $\boldsymbol{z}_{2}=(x_{2},y_{2})^T$ in $\mathbb{H}$ is given as: 
\begin{equation}
    d_{H}(\boldsymbol{z}_{1},\boldsymbol{z}_{2}) = \mathrm{arcosh}\left[1+2\frac{||\boldsymbol{z}_{1} - \boldsymbol{z}_{2}||^2}{(1-||\boldsymbol{z}_{1}||^2)(1-||\boldsymbol{z}_{2}||^2)}\right],
     \label{Rao}
\end{equation}
where $\mathrm{arcosh}$ denotes the inverse hyperbolic cosine and $\|\cdot\|$ is the usual Euclidean norm. Different from Euclidean distance, hyperbolic distance grows exponentially fast as we move the points toward the boundary of the open unit disk. There exists many models of hyperbolic geometry (Klein non-conformal model, Beltrami hemisphere model and Lorentz hyperboloid model among others). We can transform one model of hyperbolic geometry into another one by using a one-to-one mapping, which yields an isometric embedding\cite{Nielsen:2010aa}.

\subsection{Riemannian Prior on the Univariate Normal Model}\label{Riemannian}
A Gaussian distribution in $\mathbb{H}$, denoted as $\mathcal{N}_{H}(\mu,\sigma)$, depends on two parameters, the Fréchet mean $\mu\in\mathbb{H}$ (i.e., the center of mass) and the dispersion parameter  $\sigma>0$, similarly to the Gaussian density in the Euclidean space. The Gaussian probability density function in the hyperbolic space, denoted as $p_{H}(x|\mu,\sigma)$ is defined as follows \cite{Said:2014aa}:
\begin{equation}
    p_{H}(x|\mu,\sigma)=\frac{1}{Z(\sigma)}\exp\left[-\frac{d_{H}^{2}(x,\mu)}{2\sigma^{2}}\right].
    \label{pdf}
\end{equation}
Several remarks can be made from Eq. \eqref{pdf}: (i.) the main difference between the hyperbolic density $p_H(\cdot)$ and the Gaussian density in the Euclidean space is the use of the squared distance $d_{H}^{2}(x,\mu)$ in the exponential (referred to as Rao distance) and a different dispersion dependent normalization constant $Z(\sigma)$ which reduces to $\sqrt{2\pi\sigma^{2}}$ in the Euclidean case. Note that the constant $Z(\sigma)$ is linked to the underlying geometry of the hyperbolic space (ii.). To define a Gaussian distribution $\mathcal{N}_{H}(\mu,\sigma)$, through its probability density function, it is necessary to have an exact expression of the normalizing constant $Z(\sigma)$. This constant can be determined using hyperbolic polar coordinates $r=d_{H}(x,\mu)$ (i.e, a pulling-back) to calculate $Z(\sigma)$ using an integral depending on the Riemannian volume element (iii.). By introducing the parametrization $\boldsymbol{z}=(x,y)^T$ where $x=\mu/\sqrt{2}$, $y=\sigma$ and the Riemannian metric for the univariate Gaussian model is $ds^{2}(z)=(dx^{2}+dy^{2})/y^{2}$, the Riemannian area (since $\mathbb{H}$ is of dimension 2) is  $dA(z)=dxdy/y^{2}$ or $dA(z)=\sinh(r)drd\varphi$ in polar coordinates. For a two-dimensional parameter space, the normalization constant $Z(\sigma)$ was computed in \cite{Said:2014aa}, leading to:
\begin{equation}
Z(\sigma)=\int_{\mathbb{H}}^{}\exp\Big(-\frac{r^{2}}{2\sigma^{2}}\Big)dA(z)=2\pi \sigma\sqrt{\frac{\pi}{2}} \exp\left(\frac{\sigma^{2}}{2}\right)\textrm{erf}\left(\frac{\sigma}{\sqrt{2}}\right),
    \label{normal_const}
\end{equation}
where \textrm{erf} is the error function. Formula \ref{normal_const} completes the definition of the Gaussian distribution $\mathcal{N}_{H}(\mu,\sigma)$. 
In \cite{Mathieu:2019aa} authors shown that when $\sigma$ get smaller (resp. bigger), the Riemmannian normal pdf get closer (resp. futher) to the wrapped normal pdf \cite{Mathieu:2019aa}.

\section{Hyperbolic 2-space Local Outlier Probability}\label{Hyperbolic}

\begin{figure}
    \begin{minipage}[t]{.33\textwidth}
       \centering
        \includegraphics[width=\textwidth]{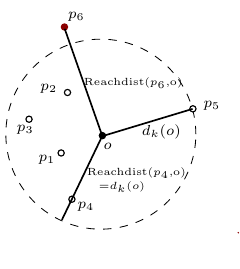}
        \subcaption{Example of reachability distance.}\label{fig:Reachability}
    \end{minipage}
  \hspace{1.2cm}
    \begin{minipage}[t]{.46\textwidth}
        \centering
        \includegraphics[width=\textwidth]{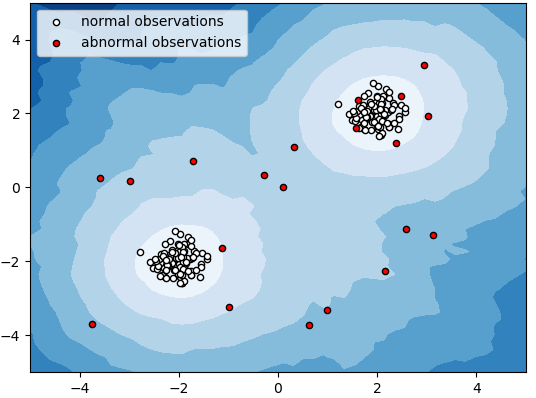}
        \subcaption{Example of LOF algorithm (scikit-learn library).}\label{fig:LOF}
    \end{minipage}  
    \label{fig:1-2}
    \caption{(a) Illustration of the reachability distance for different data points $p$ with regard to $o$, when $k=5$ . (b) The local density deviation of a given data point with respect to its neighbors.}
\end{figure}
\subsection{Density-based outlier scoring with a probabilistic approach}
This subsection briefly presents the main theoretical principles of some research studies dealing with local outlier probability concepts.
A local outlier is a data point that is different or far from most elements of the entire dataset as compared to its local neighborhood, which is measured by the k-Nearest Neighbors (kNN) algorithm \cite{Cover:1967aa}. Therefore, the local outlier detection covers a small subset of data points at a given time (Figure \ref{fig:Reachability}).
To compute the degree of outlier of a point $p$ in a dataset $\mathcal{D}$, several distances have to be introduced \cite{Hu:2016aa}. The $k$-distance of a point $p \in \mathcal{D}$ denoted as $d_k(p)$ is the distance between $p \in \mathcal{D}$ and its $k$-nearest neighbor. The notion of $k$-distance must be used to delimit a neighborhood that contains the $k$-nearest neighborhood of $p$. This neighborhood denoted as $N_k(p)$ is defined as $ N_k(p) = \left\{q \in \mathcal{D} \backslash \{p\} | d(q,p) \; \le \; d_k(p)\right\}$. Thus, the reachability distance denoted $\mathrm{reach\-dist}_k(p,o)$ of a point $p \in \mathcal{D}$ regarding a point $o$ is defined as  $\mathrm{reach\-dist}_k(p,o) = \mathrm{max}\left\{d_k(p), d(p,o)\right\}$. Based on these definitions, the LOF (Local Outlier Factor) algorithm has been introduced in \cite{Breunig:2000aa} to improve the kNN approach in the senorio where, for intance in two-dimentional data set, the density of one cluster is significantly higher (resp. lower) than another cluster. To do this, it calculates the local reachable density of the data, and calculates the local outlier factor score according to the local reachable density. Figure \ref{fig:LOF} presents the local density deviation of a given data point with respect to its neighbors. It considers as outlier samples that have a substantially lower density than their neighbors. In this work, we introduce the HLOF (Hyperbolic Local Outlier Factor)  algorithm by replacing $d(q,p)$ with the Rao distance using Eq.\eqref{Rao}. 

While the LOF (Local Outlier Factor) algorithm detects the outlier data points using the reachibility distance, the Local Outlier Probability (LoOP) algorithm introduces the probabilistic distance of $o \in \mathcal{D}$ to a context set $\mathcal{S}\subseteq \mathcal{D}$, referred to as $\mathrm{pdist}(o,\mathcal{S})$ with the following property:
\begin{equation}
\forall s\in\mathcal{S}:\mathcal{P}[d(o,s)\leq\mathrm{pdist}(o,\mathcal{S})]\geq\varphi.
\end{equation}
This probabilistic distance corresponds to the radius of a disk that contains a data point of $\mathcal{S}$, obtained from the kNN algorithm, with a certain probability, denoted as $\varphi$. The reciprocal of the probabilistic distance can be considered as an estimation of the density of $\mathcal{S}$, i.e,
 $\mathrm{pden}(\mathcal{S})=\frac{1}{\mathrm{pdist}(o,\mathcal{S})}$.
Assuming that $o$ is the center of $\mathcal{S}$ and the local density is approximately a \textit{half-Gaussian} distribution, the probabilistic set distance of $o$ to $\mathcal{S}$ can be defined as 
\begin{equation}
    \mathrm{pdist}(o, \mathcal{S}) = \lambda \sigma(o,\mathcal{S}),
    \label{pdist}
\end{equation}
where $\sigma(o, \mathcal{S}) = (\sum_{s \in \mathcal{S}}d(o, s)^2/\left|\mathcal{S}\right|)^{1/2}$ is the standard Euclidean distance of $o$ in $\mathcal{S}$ which is similar to the standard deviation. The parameter $\lambda$ is linked to the selectivity of the detection through the \textit{quantile function} of the normal distribution via the relation $\lambda=\sqrt{2} \mathrm{erfinv}(\varphi)$, where $\mathrm{erfinv}$ is the inverse error function. 

To be detected as an anomaly for the set $\mathcal{S}$, a data point should deviate from the center of $\mathcal{S}$ for more than $\lambda$ times the standard distance. For instance, $\lambda = 3$ means that a circle of radius $\text{pdist}(o, \mathcal{S})$ and center $o$ contains any data point of $\mathcal{S}$ with a probability $\varphi \approx 99.7\%$. The resulting probability is the Local Outlier Probability (LoOP) given by
\begin{equation}
    \mathrm{LoOP}_{\mathcal{S}}(o)= \mathrm{max}\left\{ 0, \mathrm{erf}\left(\frac{\mathrm{PLOF}_{\lambda,\mathcal{S}}(o)}{\mathrm{nPLOF}\sqrt2}\right)\right\},
    \label{LoOP}
\end{equation}
where the Probabilisitic Local Outlier Factor (PLOF) is defined as
$\mathrm{PLOF}_{\lambda,\mathcal{S}}(o) = (\mathrm{pdist}(\lambda,o, \mathcal{S}))/(\mathbf{\mathrm{E}}_{s\in\mathcal{S}}\left[\mathrm{pdist}(\lambda,s, \mathcal{S})\right])-1$ and a normalization factor $\mathrm{nPLOF}$ is such that
 $\mathrm{nPLOF} = \lambda (\mathrm{E}[\mathrm{PLOF}^2])^{1/2}$. The LoOP value is directly interpretable as the probability of $o$ being an outlier, i.e, close to $0$ for points within dense regions and close to $1$ for density-based outliers.  

\subsection{HLoOP Algorithm}
This subsection presents the main contribution of this study, which is an adaptation of the LoOP algorithm to data lying in a hyperbolic 2-space. 
As mentioned above, the LoOP algorithm in an Euclidean space exploits a probabilistic set distance, called $\mathrm{pdist}(o, \mathcal{S})$ (see Eq. \ref{pdist}), to pick the density around $o$ in the context set $\mathcal{S}$ with a probability of $\varphi$. The parameter $\lambda$ gives control over the approximation of the density.
To define a local outlier probability adapted to hyperbolic geometry, it is necessary to calculate a new parameter $\lambda_H$, which ensures that $\mathrm{pdist}(o, \mathcal{S})$ is performed without undermining hyperbolic geometry. 
To come up with such a solution, the key idea is to derive a new \textit{quantile function} through an expression of a Gaussian cumulative distribution function (c.d.f) that can be obtained by integrating the probability density function \eqref{pdf} in $\mathbb{H}$. Using polar coordinates (see subsection \ref{Riemannian}, remark iii.), it is possible to calculate this Gaussian c.d.f explicitly. To find the parameter $\lambda_H$, we consider the probabilistic distance of $o \in \mathcal{D}$ to a context set $\mathcal{S}\subseteq \mathcal{D}$ using a Riemannian distance $d_{H}(o,s)$ and the following statistical property:
\begin{align}
\forall s\in\mathcal{S}:\varphi=\mathcal{P}[0<d_{H}(o,s)\leq\lambda_H \sigma_{r}]=\mathcal{G}_{H}(\lambda_H \sigma_{r}). 
\label{stat}
\end{align}
Assuming that $o$ is the center of $\mathcal{S}$ and the set of distances of $s\in\mathcal{S}$ to $o$ is approximately \textit{half-Gaussian} in a hyperbolic space, one can compute the standard deviation $\sigma_{r}$ using the Riemannian distance $d_{H}(o,s)$ with a mean $d_{H}(o,o)=0$. Note that the standard deviation of $r$ denoted as $\sigma_{r}$ and its probability density function can be determined from the function $\mathcal{G}_{H}(R)=\mathcal{P}[0<r<R]$, e.g., $p_{H}(r,\sigma_{r})=\mathcal{G}_{H}'(r,\sigma_{r})$. Theorem $1$ presents the main result of this paper.

\begin{theorem}
Given $r\in\mathbb{H}$, $\sigma_{r}>0$, the Riemannian geometry of the Gaussian cumulative model associated with the distribution defined in Eq. \eqref{pdf} is given by 
\begin{equation}
\displaystyle \mathcal{G}_{H}(r,\sigma_{r})=\frac{\pi\sqrt{2\pi}\sigma_{r} \displaystyle e^{\frac{\sigma_{r}^{2}}{2}}}{2 Z(\sigma_{r})}\times
\begin{pmatrix}
\displaystyle
2 \mathrm{erf}\Big(\frac{\sigma_{r}}{\sqrt{2}}\Big)+\mathrm{erf}\Big(\frac{r-\sigma_{r}^{2}}{\sigma_{r}\sqrt{2}}\Big)-\mathrm{erf}\Big(\frac{r+\sigma_{r}^{2}}{\sigma_{r}\sqrt{2}}\Big)   
\end{pmatrix}.
\end{equation}
\end{theorem}
\begin{proof}
Let $\mathcal{P}[0<r\leq R]$ and $dA(z)=\sinh(r)drd\varphi$  such that:
\begin{align*}
\mathcal{G}_{H}(R)&=\displaystyle{\int_{0}^{2\pi}\int_{0}^{R}}\frac{1}{Z(\sigma_{r})}\exp\Big(-\frac{r^{2}}{2\sigma_{r}^{2}}\Big)\sinh(r)drd\varphi.
\end{align*}
The probability density function (pdf) $p_H(\cdot)$ must satisfy the following condition:
\begin{align}
    \int_{\mathbb{H}}^{}p_{H}(x|\mu,\sigma)d(\mu,\sigma)=1,
    \label{normalize}
\end{align}
where $d(\mu,\sigma)$ is the Lebesgue measure.
The cumulative distribution function of the univariate Gaussian distribution of pdf $p_H(\cdot)$ can be computed using Eq. \eqref{normalize} as follows:
\begin{align}
\mathcal{G}_{H}(R)&=\frac{2\pi}{Z(\sigma_{r})}\int_{0}^{R}\frac{e^{\frac{\sigma_{r}^{2}}{2}}}{2}
\begin{pmatrix}
e^{-\displaystyle\frac{(r-\sigma_{r}^{2})^{2}}{2\sigma_{r}^{2}}}-e^{-\displaystyle\frac{(r+\sigma_{r}^{2})^{2}}{2\sigma_{r}^{2}}}\end{pmatrix}dr\nonumber
\\
&=\frac{\pi\sqrt{2\pi}\sigma_{r} e^{\frac{\sigma_{r}^{2}}{2}}}{2 Z(\sigma_{r})}\begin{pmatrix}\displaystyle\frac{2}{\sqrt{\pi}}\int_{\frac{-\sigma_{r}}{\sqrt{2}}}^{\frac{R-\sigma_{r}^{2}}{\sqrt{2}\sigma_{r}}}e^{-u_{1}^{2}} du_{1}-\frac{2}{\sqrt{\pi}}\int_{\frac{\sigma_{r}}{\sqrt{2}}}^{\frac{R+\sigma_{r}^{2}}{\sqrt{2}\sigma_{r}}}e^{-u_{2}^{2}} du_{2}\end{pmatrix}\nonumber
\\
&=\frac{\pi\sqrt{2\pi}\sigma_{r} e^{\frac{\sigma_{r}^{2}}{2}}}{2 Z(\sigma_{r})}
\begin{pmatrix}
\displaystyle
2 \mathrm{erf}\Big(\frac{\sigma_{r}}{\sqrt{2}}\Big)+\mathrm{erf}\Big(\frac{R-\sigma_{r}^{2}}{\sigma_{r}\sqrt{2}}\Big)-\mathrm{erf}\Big(\frac{R+\sigma_{r}^{2}}{\sigma_{r}\sqrt{2}}\Big) 
\label{pro}
\end{pmatrix}.
\end{align}
Taking the limit $\mathcal{G}_{H}(R) \underset{R\to 1}{\longrightarrow} 1$ yields in Eq. \eqref{pro}
\begin{align*}
Z(\sigma_{r})=(2\pi \sigma_{r})\sqrt{\frac{\pi}{2}} \exp\left(\frac{\sigma_{r}^{2}}{2}\right)\textrm{erf}\left(\frac{\sigma_{r}}{\sqrt{2}}\right).
\end{align*}
We recover the formula given in \cite{Said:2014aa}, 
which completes the proof.
\end{proof}
Combining all these results, the parameter $\lambda_{H}(\sigma_{r})$ is determined by the inverse of $\mathcal{G}_{H}(\lambda_H \sigma_{r})$ (see eq.\ref{stat}) such that 
\begin{align}
\lambda_H(\sigma_{r})=\frac{1}{\sigma_{r}}\mathcal{G}_{H}^{-1}(\varphi).
\label{lambda}
\end{align}
Hence, whilethe traditional \textit{quantile function} is independent of the standard deviation we have obtained means to directly derive the parameter $\lambda_H$  that exploits the underlying geometry of the hyperbolic space (see subsection \ref{Riemannian}, remark ii.). The HLoOP algorithm is summarized in Algorithm~$1$.
\begin{algorithm}
\caption{The procedure of HLoOP algorithm}\label{alg:cap}
\renewcommand{\algorithmicrequire}{\textbf{Input:}}
\renewcommand{\algorithmicensure}{\textbf{Output:}}
\begin{algorithmic}
\Require The data set $X=\{x^{i}\}_{i=1}^{m}$ where $x^{i}=(x_{1}^{i},x_{2}^{i},\cdots,x_{n}^{i})$ $\in \mathbb{R}^{n}$.\\
\hspace{0.8cm}Pre-determined threshold $\varphi$, parameter $k$ and hyperbolic distance $d_{H}(p,q)$;
\vspace{0.2cm}
\begin{enumerate}[label={(\arabic*)}]
    \item \textbf{Determine} the context set $\mathcal{S}$ of the data point $x^{i}$ from kNN algorithm;
    \item \textbf{Compute} the standard distance $\sigma_r$ of the context set $\mathcal{S}$;
    \item \textbf{Determine} $\mathcal{G}_{H}^{-1}(\varphi)$ to derive the parameter $\lambda_{H}$ by eq.\ref{lambda};
    \item \textbf{Calculate} the probabilistic set distance $\mathrm{pdist}_{k}(x^{i})$ of the data point $x^{i}$ by eq.\ref{pdist}; 
    \item \textbf{Compute} the local outlier probability $\mathrm{LoOP}_{k}(x^{i})$ of the data point $x^{i}$ by eq.\ref{LoOP}; 
\end{enumerate}
\vspace{0.2cm}
\Ensure Abnormal data points.
\end{algorithmic}
\end{algorithm}

\newpage
\subsection{Implementation Details}
\begin{figure}[htbp]
    \begin{minipage}[t]{.49\textwidth}
       \centering
        \includegraphics[width=\textwidth]{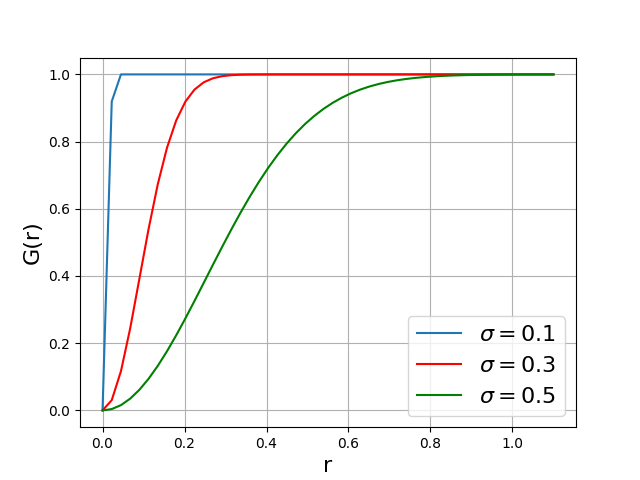}
        \subcaption{\centering Cumulative univariate Gaussian model associated with the distribution Eq. \eqref{pdf}.}\label{fig:step_1_ginv}
    \end{minipage}
    \begin{minipage}[t]{.49\textwidth}
        \centering
        \includegraphics[width=\textwidth]{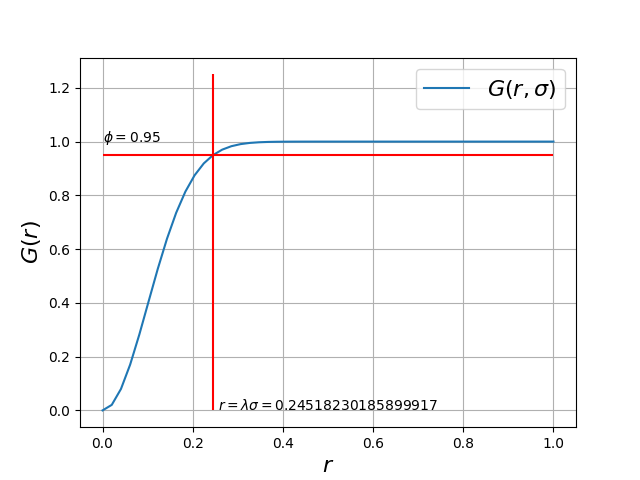}
        \subcaption{\centering Newton's method to determine $r = \mathcal{G}^{-1}(\varphi)$.}\label{fig:Ginv_step2}
    \end{minipage}  
    \label{fig:1-3}
    \caption{Cumulative distribution $\mathcal{G}_H$ in the Poincare disk and resolution of $\mathcal{G}_H(r, \sigma) = \varphi$.}
\end{figure}

Before presenting the performance of the algorithm described above, it is interesting to discuss some aspects related to the implementation of the HLoOP algorithm. Most of the steps used in the implementation of the HLoOP algorithm are directly related to the Euclidean LoOP, except that the distances are no longer Euclidean but hyperbolic. However, the computation of the significance $\lambda$ cannot be computed as in the Euclidean LoOP. While an analytic expression of the Gaussian quantile function is known in the Euclidean space, the derivation of the cumulative distribution in the Poincare disk, illustrated on Figure \ref{fig:step_1_ginv}, does not lead to an analytic formulation of its inverse $\mathcal{G}_H^{-1}$. Actually, an analytical expression of $\mathcal{G}_H^{-1}$ is not needed to compute $\lambda$ providing the value of $r = \lambda \sigma$ for which $ \mathcal{G}_H(r, \sigma) = \varphi$ can be determined. This is equivalent to solving:
\begin{align*}
    \mathcal{G}_H(r, \sigma)  = \varphi, 
\end{align*}
for a given pair $(\sigma, \varphi)$, which can be done using Newton's method. Once we have obtained $r$, as shown on Figure \ref{fig:Ginv_step2}, the significance can be determined by using the relation $r = \lambda \sigma$, yielding $\lambda = r/\sigma$.
All the elements required to implement the HLoOP algorithm are now available. The next Section is dedicated to the assessment of the performance of this algorithm.

\section{Results}\label{Results}

\subsection{Performance of the HLoOP algorithm on a toy dataset}
\begin{figure}[htbp]
    \centering
    \includegraphics[scale = 0.38]{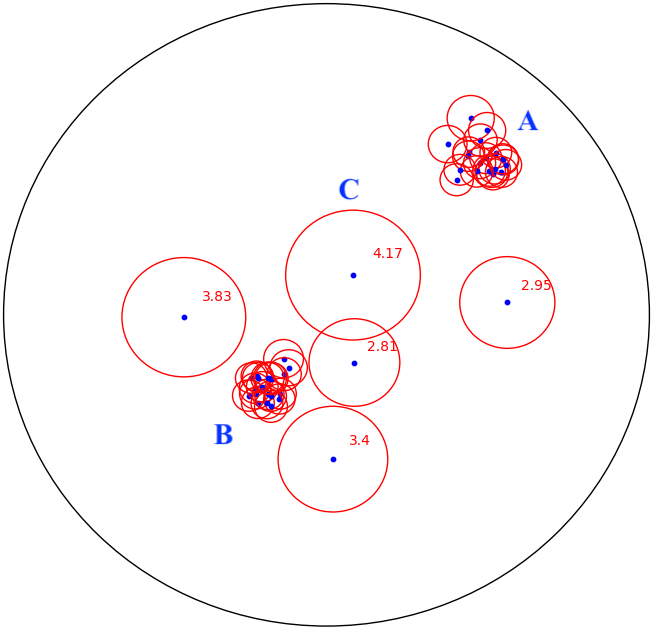}
    \caption{\centering Embedding of the toy dataset in the Poincaré disk : HLOF values.}\label{fig:embedding_toy_dataset_HLOF}
\end{figure}

\begin{figure}[htbp]
    \centering
    \includegraphics[scale = 0.38]{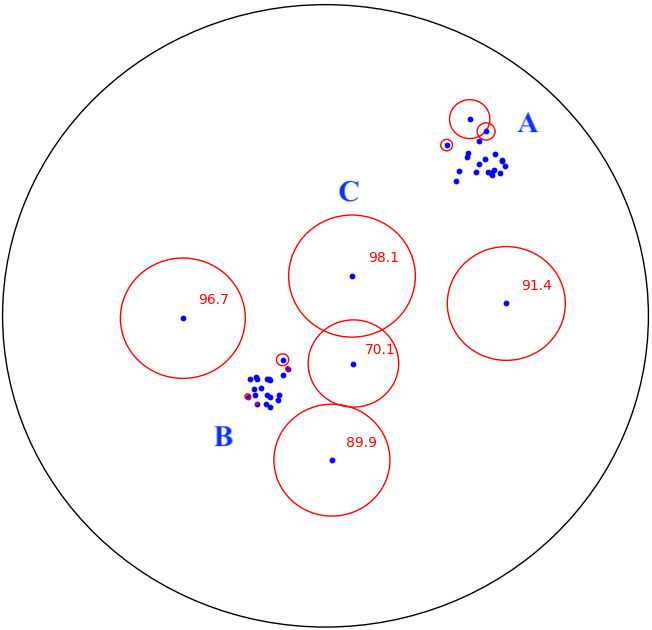}
    \caption{\centering Embedding of the toy dataset in the Poincaré disk : HLoOP values -- Some points have very small H-LoOP values and their associated circle do not appear in the figure --.}\label{fig:embedding_toy_dataset_HLoOP}
\end{figure}

The HLoOP method is first used to detect the outliers in a toy dataset, with a reduced number of points. The dataset was generated as follows: first, some vectors were generated uniformly in two circular areas 
located in the Poincaré Disk (clusters \textbf{A} and \textbf{B}). Then, each area is filled with 40 points whose positions are pulled from the normal distribution $\mathcal{N}(\cdot,R \boldsymbol{I}_2)$ where $\boldsymbol{I}_2$ is the $2 \times 2$ identity matrix. 
Five points located outside these areas (cluster \textbf{C}) constitute the outliers of the toy dataset, which is finally composed of $2 \times 40 + 5 = 85$ points. The HLoOP algorithm is applied to this dataset and its performance is compared to that of HLOF.
As a first test, we compute the HLOF and HLoOP values of each point of the embedding for $k = 15$ and, for HLoOP a treshold $\varphi = 95\%$. Figures \ref{fig:embedding_toy_dataset_HLOF} and \ref{fig:embedding_toy_dataset_HLoOP} show the different points that are surrounded by a circle whose radius is proportional to the HLoOP or HLOF value. We observe that for both methods (HLoOP or HLOF), the outliers (cluster \textbf{C}) have a score higher than the inliers (clusters \textbf{A} and \textbf{B}). For the HLoOP, this correspond to the probability of a point to be an outlier, while for the HLOF, the interpretation of the score is less straightforward. It is also interesting to note that cluster \textbf{A} highlights a weakness of HLOF : like LOF, it is designed for clusters of uniform density. The probability of datapoints in cluster \textbf{A} being generated by Gaussian distribution, the HLOF assigns high outlier scores while these points were in fact generated by the cluster. The HLoOP value is much more useful here : there is a clear chance the the point is an outlier, but it is also very likely it is just an outer point of the clusters normal distribution. \newline 

\newpage

\begin{figure}[h]
    \centering
    \includegraphics[scale = 0.6]{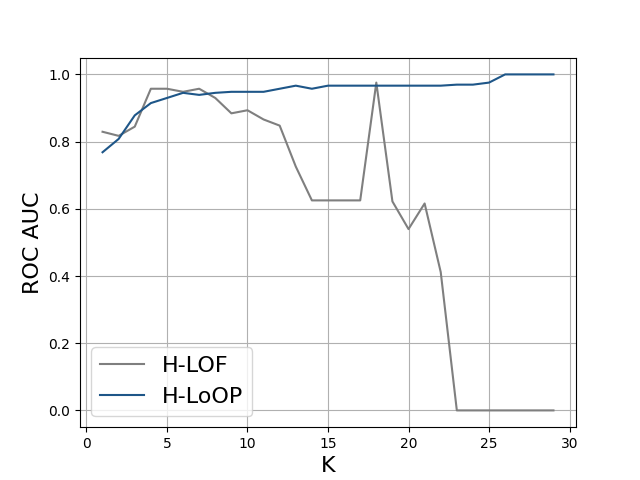}
    \caption{AUC ROC - Outlier detection in a toy dataset}
    \label{fig:auc_roc_toy_dataset}
\end{figure}
The metric used to quantify the quality of the outlier detection is the Area Under the Receiving Operator Curve (AUC-ROC). We recall that  value of AUC-ROC near 0 coreesponds to very poor detection performance (near 0\% of the decision made by the algorithm are correct), while an AUC ROC close to 1 means that the algorithm is making very few errors. As observed in Figure \ref{fig:auc_roc_toy_dataset}, the HLoOP algorithm provides very good anomaly detections: for $k>2$ (number of neighbors considered to evaluate the density of the context set $\mathcal{S}$), the number of true positives (actual outliers detected as outliers) is between 95 and 100 \%, which is a very good result. In the meantime, the performance of HLOF is more contrasted and strongly dependent on the value of $k$. In particular, for higher values of $k$, the HLOF performance dramatically decreases. With such promising results, the next section aims at assessing the performances of HLoOP on a bigger dataset, containing up to 1000 points. 

\subsection{Evaluating the performances on the Wordnet/Mammals subgraph}
\begin{figure}[h]
    \centering
    \includegraphics[scale=0.6]{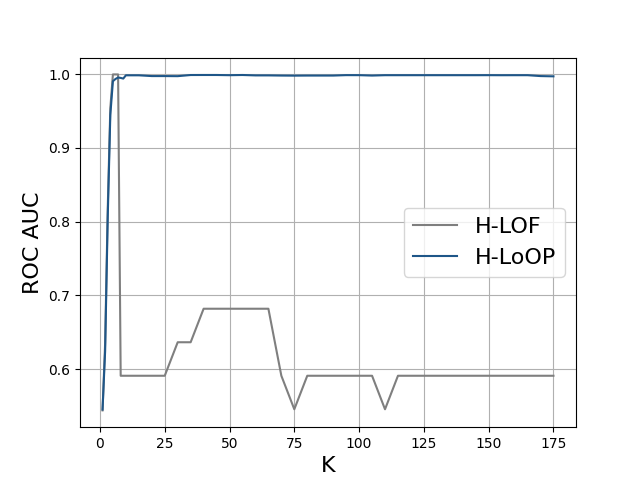}
    \caption{AUC ROC - Outlier detection in our corrupted subgraph of Wornet/Mammals}
    \label{fig:auc_roc_wordnet}
\end{figure}

This section evaluates the performance of H-LoOP on a subgraph of the WORDNET database. WORDNET is a lexical dataset composed by 117000 synsets, which corresponds to nouns, adjectives or verbs that are linked by conceptual relations. Several subgraphes are known to exist in this dataset. Among them, we decided to apply H-LOF and H-LoOP on a group of 1180 synsets from the subgraph ``Mammals''. The dataset was corrupted by $11$ outliers corresponding to nouns of animals that are not mammals (i.e., fishes, reptiles or birds) and was embedded in the Poincaré Disk using the algorithm of Nickel et al. (2017). The values of HLoOP and HLOF were calculated for the points of this embedding. The Area Under the Receiving Operator Curve (AUC ROC) was finally calculated for both HLOF and HLoOP for several value of $K$. As shown in Figure \ref{fig:auc_roc_wordnet}, the performance of HLoOP is better than HLOF  for all values of $K$, with a ROC AUC larger than 0.98, while the HLOF leads to a ROC value less than 68\%. In addition to its good performance, the HLoOP algorithm leads to AUC values that are quite independent of $K$, which is outstanding.

\section{Conclusion and perspectives}

This paper has presented extensions of the Local Outlier Factor (LOF) and Local Outlier Probability (LoOP) algorithms, respectively refered to as Hyperbolic LOF (HLOF) and Hyperbolic LoOP (HLoOP). Rather than working in the Euclidean space, these extensions work in a specific model of the hyperbolic space, namely the Poincaré Disk. Both algorithms are density based and compare the density of a point's neighborhood with the density of others's neighborhood. On one hand, the HLOF compute the density based on a deterministic distance (called reachibility distance) while the HLoOP introduces the notion of probabilistic distance and returns for each point its probability of outlierness. Simulations results conducted on a toy dataset have shown that the HLoOP algorithm allows a better distinction of outliers and inliers when compared to HLOF. While the HLoOP directly provides the probability of each point to be an outlier, the HLOF returns a score whose interpretation is not straightforward and depends on the dataset under study. Evaluations of the areas under the receiver operational characteristics of data in the Poincaré disk have confirmed a better detection performance of HLoOP when compared to HLOF. The results obtained with this dataset have also shown that the HLoOP performance seems to be less sensitive to the number of neighbors taken into account in the computation of the density of the context set than for HLOF. Given these promising results, we have embedded the \textit{mammals} subset of the Wordnet dataset in the Poincaré disk after introducing artificial outliers. The HLOF and HLoOP values and the areas under the receiver operational characteristics HLOF and HLoOP algorithms confirm the results obtained with the previous dataset. \newline 
Future work includes the extension of these two algorithms to the Lorentz's disk, i.e., to another model of the hyperbolic space. Indeed, it has been shown that the Poincaré disk presents some numerical instabilities that are not observed in the Lorentz model. Moreover, it would be interesting to apply the HLoOP and HLOF algorithms to more complex datasets, with more points and more attributes. For instance, given the growing interest of the hyperbolic geometry in the computer vision domain, it could be worthy to try using the HLoOP and HLOF to detect outliers in a set of images. Finally, the hyperbolic geometry could be used to derive new outlier detection algorithms based on isolation forest or one-class support vector machines. 

\bibliography{biblio} 
\bibliographystyle{achemso}

\end{document}